\documentclass[letterpaper]{article} 
\usepackage{aaai2026}  
\usepackage{times}  
\usepackage{helvet}  
\usepackage{courier}  
\usepackage[hyphens]{url}  
\usepackage{graphicx} 
\usepackage{natbib}  
\usepackage{caption} 
\usepackage{algorithm}
\usepackage{algorithmic}

\usepackage{amsmath}
\usepackage{amssymb}
\usepackage{multirow}
\usepackage{booktabs}
\usepackage{colortbl}
\usepackage{xcolor}
\usepackage{array}
\usepackage{pifont}
\usepackage{bm}
\usepackage[misc]{ifsym}

\definecolor{mycolor}{RGB}{241,240,255}

\usepackage{newfloat}
\usepackage{listings}
\DeclareCaptionStyle{ruled}{labelfont=normalfont,labelsep=colon,strut=off} 
\floatstyle{ruled}
\newfloat{listing}{tb}{lst}{}
\floatname{listing}{Listing}
\title{CASL: Curvature-Augmented Self-supervised Learning for 3D Anomaly Detection}
\author{
    Yaohua Zha\textsuperscript{\rm 1,2}, \ 
    Xue Yuerong\textsuperscript{\rm 1}, \ 
    Chunlin Fan\textsuperscript{\rm 1}, \ 
    Yuansong Wang\textsuperscript{\rm 1}, \  
    Tao Dai\textsuperscript{\rm 3}\thanks{Corresponding author. (daitao.edu@gmail.com)}, \\
    Ke Chen\textsuperscript{\rm 2}, \ 
    Shu-Tao Xia\textsuperscript{\rm 1,2} \\
}
\affiliations{
    \textsuperscript{\rm 1}Tsinghua Shenzhen International Graduate School, Tsinghua University\\
    \textsuperscript{\rm 2}Institute of Perceptual Intelligence, Pengcheng Laboratory\\
    \textsuperscript{\rm 3}College of Computer Science and Software Engineering, Shenzhen University\\
}

\usepackage{bibentry}

\begin{document}

\maketitle

\begin{abstract}

Deep learning-based 3D anomaly detection methods have demonstrated significant potential in industrial manufacturing. However, many approaches are specifically designed for anomaly detection tasks, which limits their generalizability to other 3D understanding tasks. In contrast, self-supervised point cloud models aim for general-purpose representation learning, yet our investigation reveals that these classical models are suboptimal at anomaly detection under the unified fine-tuning paradigm. 
This motivates us to develop a more generalizable 3D model that can effectively detect anomalies without relying on task-specific designs.
Interestingly, we find that using only the curvature of each point as its anomaly score already outperforms several classical self-supervised and dedicated anomaly detection models, highlighting the critical role of \textbf{curvature} in 3D anomaly detection.
In this paper, we propose a \textbf{C}urvature-\textbf{A}ugmented \textbf{S}elf-supervised \textbf{L}earning (\textbf{CASL}) framework based on a reconstruction paradigm. Built upon the classical U-Net architecture, our approach introduces multi-scale curvature prompts to guide the decoder in predicting the spatial coordinates of each point. Without relying on any dedicated anomaly detection mechanisms, it achieves leading detection performance through straightforward anomaly classification fine-tuning.
Moreover, the learned representations generalize well to standard 3D understanding tasks such as point cloud classification. The code is available at \url{https://github.com/zyh16143998882/CASL}.
\end{abstract}

\section{Introduction}

Deep learning-based anomaly detection algorithms \cite{deepad,tsad,mvtec,iad} have played a significant role in industrial manufacturing due to their efficiency in quality control. Early research \cite{ssaead,stad,patchcore} primarily focused on the 2D domain and achieved notable progress. More recently, 3D point cloud anomaly detection \cite{real3dad,anoshapenet,group3ad} has emerged as a prominent research direction, as its ability to capture more comprehensive spatial structures leads to significantly improved detection performance compared to traditional 2D approaches.

Existing 3D anomaly detection methods can be broadly categorized into feature-matching-based methods \cite{patchcore,real3dad,group3ad} and reconstruction-based methods \cite{anoshapenet,r3dad}. Feature-matching methods build a feature pool from normal samples during training and identify anomalies at inference by computing feature distances. Reconstruction-based methods learn to reconstruct normal samples from local point information during training and detect anomalies by comparing the reconstructed data with the original input. These methods are tailored specifically for anomaly detection tasks. Despite achieving impressive detection performance, their task-specific design inherently restricts their generalizability to broader 3D understanding tasks.

\begin{figure}[t]
    \begin{center}
    \includegraphics[width=\linewidth]{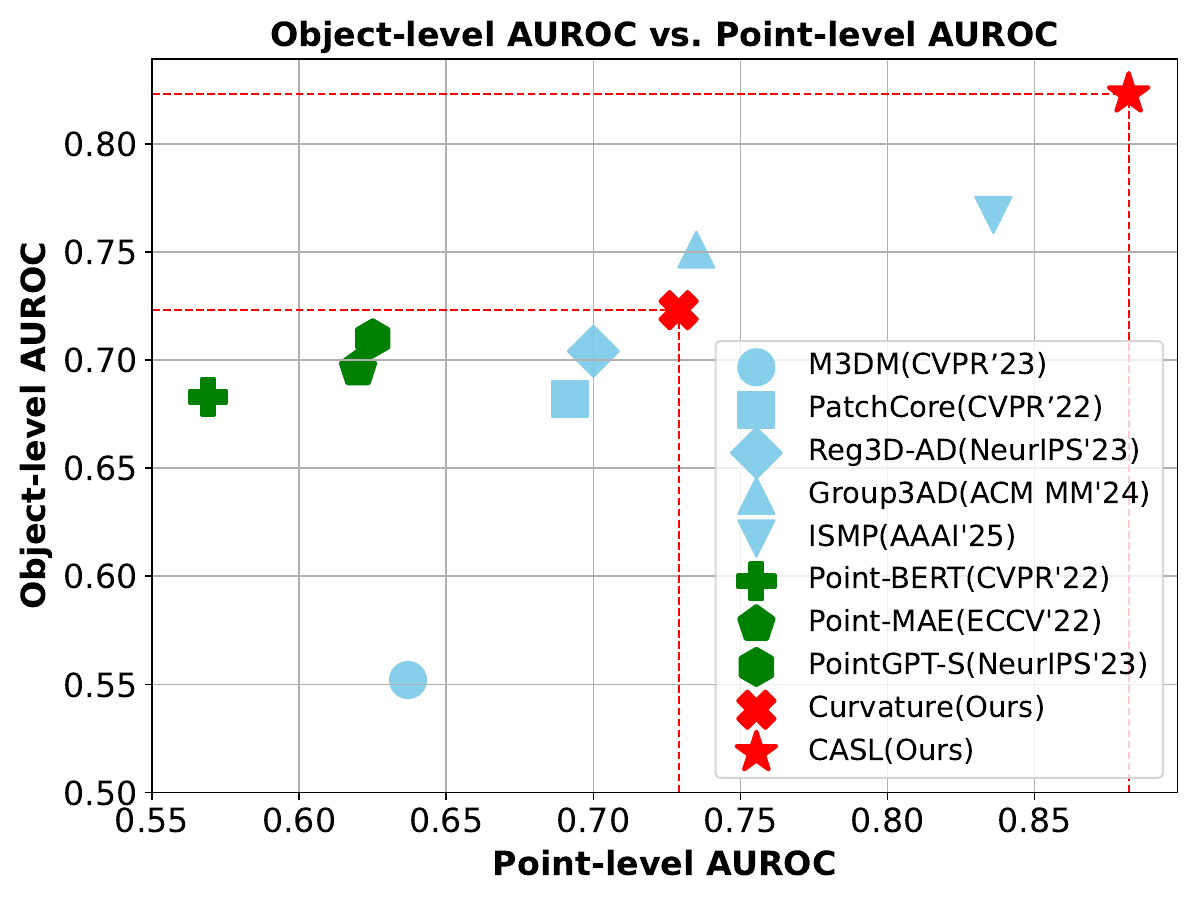}
    \caption{Anomaly detection performance of different methods on the Real3D-AD dataset. Blue markers represent task-specific anomaly detection methods, while green markers correspond to classical self-supervised models fine-tuned under the “pre-training and fine-tuning” paradigm. Red markers denote our results: Curvature refers to our non-learning-based method, and CASL represents the proposed curvature-augmented self-supervised learning framework.
    }\label{scatter}
    \end{center}
\end{figure}

In contrast, self-supervised representation learning on point clouds \cite{pointmae,idpt,pma,pointgpt} aims to learn general-purpose representations through a unified “pre-training and fine-tuning” paradigm. For instance, methods such as Point-MAE \cite{pointmae}, CrossPoint \cite{crosspoint}, and PointGPT \cite{pointgpt} first pre-train models using self-supervised signals and then transfer them to downstream tasks such as classification and segmentation via simple task-head-based fine-tuning. This unified paradigm significantly enhances the generalizability of the models. Although some previous 3D anomaly detection methods (e.g., PatchCore \cite{patchcore}, Reg3D-AD \cite{real3dad}) have employed pre-trained point cloud models, they primarily rely on feature matching mechanisms specifically designed for anomaly detection, rather than adhering to the unified “pre-training and fine-tuning” paradigm. Therefore, these methods offer limited evidence regarding the generalizability of pre-trained models to anomaly detection tasks.

To this end, we begin by fine-tuning existing pre-trained point cloud models using a simple pseudo-anomaly-based classification task, aiming to assess their generalizability to anomaly detection scenarios. As illustrated in Figure \ref{scatter}, our experimental results show that these classical models exhibit suboptimal performance when applied to anomaly detection via this straightforward fine-tuning approach. We attribute this suboptimal performance to the prevalent issue of geometric shortcuts \cite{pointcpr,pcpmae} in existing classical self-supervised methods. These approaches typically perform reconstruction directly from the coordinate space to the coordinate space, resulting in a complete overlap between the semantic and positional domains. Consequently, the learned representations overly rely on low-level spatial features, leading to fine-grained representational collapse in anomaly detection.

Furthermore, we make an interesting observation: \textit{\textbf{a non-learning-based method that uses only the curvature of each point as its anomaly score outperforms several classical self-supervised and task-specific anomaly detection models}}, as shown in Figure \ref{scatter}. This finding underscores the potential of curvature as a highly informative cue for identifying structural irregularities. As an intrinsic geometric property parallel to the coordinate domain, curvature can effectively alleviate the issue of geometric shortcuts when used as the semantic representation in self-supervised learning, offering a compelling alternative to traditional coordinate-based semantics.

In this paper, we propose a curvature-augmented representation learning framework grounded in a reconstruction paradigm. Built upon the classical U-Net \cite{unet} architecture, our method introduces multi-scale curvature prompts to guide the decoder in predicting the spatial coordinates of each point. In contrast to conventional reconstruction-based approaches, which typically mask a subset of coordinates and train the model to recover their coordinates, our framework adopts a fundamentally different strategy: we mask out all point coordinates and rely exclusively on curvature-based prompts to reconstruct the original coordinates. By embedding curvature, a geometric cue that is highly sensitive to anomalies, into the learning process, the model is encouraged to focus on local surface variations and fine-grained structural details. This design compels the network to learn rich geometric representations solely from curvature information, thereby not only avoiding geometric shortcuts but also deepening the understanding of 3D structural patterns. Remarkably, without relying on any dedicated anomaly detection mechanisms, the proposed framework achieves leading performance through simple classification-based fine-tuning, improving the average Object-level AUROC (O-AUROC) by 5.6\% on the Real3D-AD \cite{real3dad} dataset and 4.8\% on the Anomaly-ShapeNet \cite{anoshapenet} dataset. Furthermore, the learned representations also exhibit strong generalization ability across standard 3D understanding tasks.

Our main contributions are summarized as follows:

\begin{itemize}
    \item We find that a non-learning method based solely on point-wise curvature outperforms classical self-supervised and several anomaly-specific approaches on benchmark datasets, highlighting the importance of curvature in 3D anomaly detection.
    \item We develop a curvature-augmented representation learning framework (CASL) based on a reconstruction paradigm. We introduce a full coordinate masking strategy combined with multi-scale curvature prompts to guide the prediction of original coordinates, effectively avoiding geometric shortcuts and promoting a deeper understanding of 3D structural patterns.
    \item Our model enables effective anomaly detection without relying on any anomaly-specific detection mechanisms, requiring only a simple pseudo-anomaly-based classification fine-tuning. It significantly outperforms existing methods on the Real3D-AD and Anomaly-ShapeNet datasets, and demonstrates superior generalization across other 3D understanding tasks.
\end{itemize}

\section{Related Work}

\subsection{Point Cloud Self-supervised Learning}

Self-supervised learning (SSL) methods for 3D point clouds can be broadly categorized into discriminative and reconstruction paradigms. Discriminative methods such as PointContrast \cite{pointcontrast} and CrossPoint \cite{crosspoint} learn representations by pulling together augmented views of the same instance while pushing apart different instances via contrastive loss. Reconstruction-based methods \cite{pointbert,pointmae,spdf,lcm,pointgpt} typically learn general-purpose 3D representations by predicting the masked regions of a point cloud from the unmasked ones, such as point-MAE \cite{pointmae} and Point-FEMAE \cite{femae} utilize autoencoders, whereas PointGPT \cite{pointgpt} employs autoregression. Some cross-modal work \cite{act,recon,i2pmae} enhance 3D representations by incorporating vision and language priors. Point-MoDE \cite{pointmode} is the first to propose complementary learning between the scene and object domains in 3D space. These methods have achieved some progress across classical 3D tasks. However, these methods commonly suffer from geometric shortcut issues, leading to representational collapse in fine-grained anomaly detection. 

\subsection{3D Anomaly Detection}

3D anomaly detection methods can be broadly classified into feature-matching-based methods and reconstruction-based methods. Feature-matching-based methods, such as Reg3D-AD \cite{real3dad} and Group3AD \cite{group3ad}, construct memory banks using features extracted from pre-trained models, comparing test samples to detect deviations. Some methods, including CPMF \cite{cpmf} and PointAD \cite{pointad}, utilize 2D projections and vision-language models for memory construction or zero-shot detection. Reconstruction-based methods, on the other hand, aim to reconstruct normal point clouds and identify anomalies through reconstruction error. IMRNet \cite{anoshapenet} improves Point-MAE \cite{pointmae} with geometric-preserving operations, while R3D-AD \cite{r3dad} employs diffusion-based reconstruction for better localization. Shape-Guided \cite{shapeguide} methods leverage dual memory banks from RGB and 3D features. Despite strong results, these methods are explicitly designed for anomaly detection. We explore a new perspective by adapting self-supervised models to perform anomaly detection under a unified “pre-training and fine-tuning” paradigm, offering a more general and flexible solution to 3D anomaly detection.

\section{Methodology}

\subsection{Problem Definition for 3D Anomaly Detection}

Given a training set $T = \{p_i \} _{i=1}^M$, where each $p_i\in \mathbb{R}^{N\times3}$ is a 3D point cloud sampled from a normal object class $c\in \mathcal{C}$, and ${C}$ denotes the set of all training classes. Here, $M$ denotes the number of point clouds in the training set, and $N$ is the number of points in each point cloud. Each training class $c$ is associated with a small number of clean samples, and no anomalous instances are available during training. At test time, given a query point cloud $x$ from a target class $c$, the objective is twofold: (1) Anomaly classification, which aims to assign a binary label $y\in \{0,1\}$, where $ y=1$ indicates that the point cloud $x$ is anomalous, and (2) Anomaly localization, which involves predicting a point-level anomaly score function $\phi: \mathbb{R}^{3} \rightarrow \mathbb{R}^{+}$, where $\phi(x_i)$denotes the likelihood of each point $x_i\in x$ being anomalous.

\subsection{Curvature Computation for Point Clouds}

Curvature is a fundamental geometric property in 3D space. Following previous practices \cite{cur1,cur2}, we can directly compute the curvature of each point from the spatial coordinates of the point cloud. Specifically, we compute point-wise curvature by analyzing the local geometric structure using the covariance matrix of neighboring points. For each point $x_i\in \mathbb{R}^{1\times3}$ , we first retrieve its $k$-nearest neighbors $\{x_{j} \} _{j=1}^k$ and compute the centroid $\overline{x_i}$ of the neighborhood. The neighborhood is then centered by subtracting the centroid from each neighbor, and a $3\times3$ covariance matrix is formed as: 
\begin{gather}
    \label{eq1}
    C_i = \frac{1}{k - 1} \sum_{j=1}^{k} (\mathbf{x}_j - \bar{\mathbf{x}}_i)(\mathbf{x}_j - \bar{\mathbf{x}}_i)^\top.
\end{gather}

We perform eigenvalue decomposition on $C_i$ to obtain the ordered eigenvalues $\lambda_1^i \leq \lambda_2^i \leq \lambda_3^i$, and define the curvature at $x_i$ as:
\begin{gather}
    \label{eq2}
    \text{Curv}(x_i) = \frac{\lambda_1^i + \lambda_2^i + \lambda_3^i}{\lambda_1^i},
\end{gather}
this normalized curvature metric captures the degree of variation in the local surface geometry and has been widely adopted as a property for point cloud analysis tasks.

\subsection{Curvature for 3D Anomaly Detection}

\begin{figure}[t]
    \begin{center}
    \includegraphics[width=\linewidth]{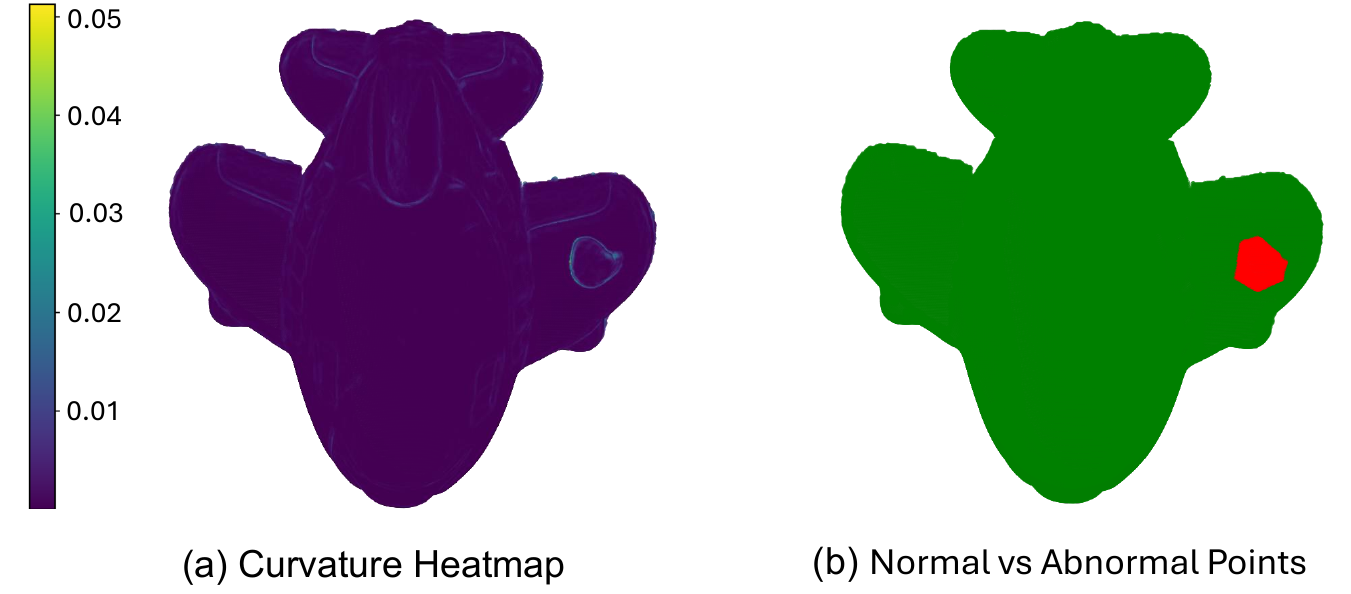}
    \caption{(a) Curvature heatmap of an anomalous point cloud, where warmer colors indicate higher curvature. (b) Spatial distribution of normal (green) and abnormal (red) points. Anomalous point boundaries often exhibit sharp increases in curvature.
    }\label{heatmap}
    \end{center}
\end{figure}

Curvature reflects the variation of surfaces in 3D space, such as concavities and convexities. We observe that anomalous points on 3D surfaces are highly correlated with the magnitude of curvature. As shown in Figure \ref{heatmap}, we present a comparison between the curvature heatmap (a) and the distribution of normal and abnormal points (b) within an anomalous point cloud. It is clearly observable that the curvature at the edges of abnormal regions is significantly higher than that of the surrounding normal regions, indicating a certain relationship between curvature and the anomaly status of the point. 

Furthermore, we make a interesting observation: by using the curvature of each point as its anomaly score, we achieve significantly better anomaly detection performance compared to several classical methods, despite this being a non-learning-based approach. As shown in Figure \ref{scatter}, the curvature-based detection method achieves significantly higher object-level AUROC and point-level AUROC on the Real3D-AD dataset compared to self-supervised approaches such as Point-BERT, Point-MAE, and PointGPT, as well as task-specific anomaly detection methods including M3DM, PatchCore, and Reg3D-AD. More quantitative results can be found in Table \ref{tabreal3d} and Table \ref{tabano}. This finding highlights the pivotal role of curvature as an intrinsic geometric property for identifying anomalies in 3D data. It suggests that curvature itself can serve as a highly informative cue for detecting structural irregularities and should be explicitly incorporated into 3D representation learning.

\begin{figure*}[t!]
    \begin{center}
    \includegraphics[width=\linewidth]{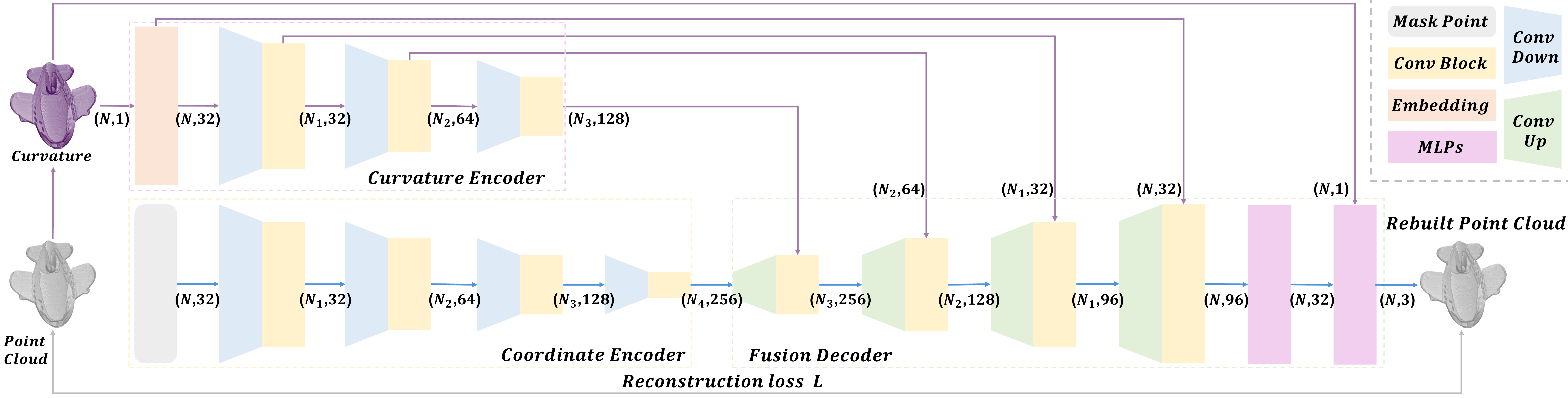}
    \caption{The pipeline of our \textbf{C}urvature-\textbf{A}ugmented \textbf{S}elf-supervised \textbf{L}earning (\textbf{CASL}) framework. Our framework, built upon the U-Net architecture, leverages multi-scale curvature prompts to guide the recovery of spatial coordinates for all masked points from random initialization.
    }\label{framework}
    \end{center}
\end{figure*}

\subsection{Curvature-Augmented Self-supervised Learning}

The overall pipeline of our curvature-augmented self-supervised learning framework is shown in Figure \ref{framework}. 

\subsubsection{Motivation for Multi-Scale Curvature Prompts.}

In point cloud reconstruction, the decoder is responsible for rebuilding spatial coordinates from high-level latent features. Providing explicit geometric priors to the decoder is crucial. Traditional approaches typically utilize the coordinates of unmasked points as geometric semantic priors to reconstruct the masked ones. While this strategy has achieved some success, it inevitably suffers from the ``\textbf{\textit{geometric shortcut}}'' issue due to the overlap between the semantic and positional domains, leading to fine-grained representational collapse in anomaly detection.

In this work, we propose addressing this issue from a curvature perspective. Curvature, as an intrinsic geometric property of 3D surfaces, plays a vital role in capturing structural irregularities. At small scales, curvature is highly sensitive to local surface variations, where subtle edges or protrusions often result in sharp curvature changes. At larger scales, curvature captures the overall shape and contour of the surface, enabling a more comprehensive understanding of global geometry. Thus, curvature serves as a meaningful geometric prior that not only provides rich structural prompts for the decoder but also avoids the geometric shortcut in traditional coordinate-to-coordinate reconstruction. To this end, we propose injecting multi-scale curvature prompts into the decoder to provide implicit geometric guidance for reconstruction. 

\subsubsection{Network Architecture and Data Pipeline.}

Given the information-intensive nature of 3D anomaly detection, where individual 3D object often comprise hundreds of thousands of points and require fine-grained, point-level anomaly localization, the classical Transformer \cite{attention} architecture is computationally inefficient. This inefficiency stems from its coarse-grained patch-based operations and quadratic time complexity of $O(n^2)$, which limit its suitability for high-resolution 3D anomaly detection. To enhance efficiency, our framework adopts the classical U-Net \cite{unet} architecture built upon Minkowski convolution \cite{mink1,mink2}, a well-established design for high-resolution point cloud processing that offers substantially greater efficiency than Transformer-based approaches.

As shown in Figure \ref{framework}, our network architecture consists of three main components: a curvature encoder, a coordinate encoder, and a fusion decoder. 

Our curvature encoder begins with an MLP-based embedding layer to extract point-wise curvature features. This is followed by three encoder blocks to capture multi-scale curvature representations. Each encoder block first applies a Minkowski Convolution (Conv Down) with a kernel size and stride of 2 to downsample the resolution, and then employs a convolutional block (Conv Block) consisting of four stride=1 convolution layers with residual connections to extract features. The detailed architecture is provided in the supplementary material.

The coordinate encoder takes as input the randomly initialized features of the $N$ masked points and passes them through four consecutive encoder blocks, progressively mapping the masked features to high-dimensional features. The final output is a tensor of shape $N_4\times256$, where $256$ denotes the feature dimension.

In our fusion decoder, the features from the previous resolution, with shape $N_{i-1}\times C_{i-1}$, is first upsampled to the current resolution $N_{i}\times C_{i-1}$ using a Minkowski Transpose Convolution (Conv Up) with both kernel size and stride set to 2. The upsampled features are then concatenated with the curvature prompts at the current resolution, resulting in a tensor of shape $N_{i}\times (C_{i-1}+C_i)$. This tensor is subsequently passed through a convolutional block (Conv Block) to produce an output of shape $N_{i}\times C_i$. This process is repeated across multiple scales until the original resolution is fully recovered.

The output of the fusion decoder is a tensor of shape $N\times 96$. We first apply MLPs to reduce its dimensionality to $N\times 32$. This representation is then concatenated with the original curvature features and passed through the second MLPs, which map it into the 3D coordinate space. In this way, the final reconstructed point cloud is obtained.

\subsubsection{Loss Function.} Due to the large number of points (typically over 100k) in the reconstructed object point clouds, classical point cloud reconstruction losses such as the Chamfer Distance (CD) \cite{cdloss} and Earth Mover's Distance (EMD) \cite{emdloss} become computationally infeasible. To ensure efficiency, we instead adopt a combination of $\ell_1$ and $\ell_2$ losses to constrain the distance between the reconstructed and ground truth point clouds. The loss is computed as follows:
\begin{gather}
    \label{eq3}
    \mathcal{L}_{{recon}} = \mathcal{L}_{{1}}(p,p_{rec}) + \mathcal{L}_{{2}}(p,p_{rec}),
\end{gather}

\subsection{Fine-tuning via Pseudo-Anomaly Classification}

The prevalent strategy in transferring pretrained models for 3D understanding tasks follows a unified “pre-training and fine-tuning” paradigm, where a pretrained backbone is adapted to downstream tasks via a lightweight task-specific head. In contrast, prior anomaly detection methods based on pretrained models often rely on reconstruction or feature matching, which are specifically tailored for anomaly detection and exhibit limited generalization to other tasks. We propose a unified fine-tuning strategy based on pseudo-anomaly classification. By synthesizing pseudo-anomalies from normal samples, our method only requires appending a binary classification head after the first MLPs in Figure \ref{framework}, thus enabling anomaly detection in a manner consistent with standard fine-tuning pipelines. Moreover, since reconstruction is not required during fine-tuning, our coordinate encoder leverages embeddings extracted directly from the coordinates rather than the randomly initialized masked points used during pre-training, enabling a more comprehensive 3D understanding.

\subsubsection{Pseudo-anomaly Generation.} The pseudo-anomaly generation strategy simulates abnormal regions by applying controlled perturbations to normal point clouds without requiring real anomaly data. We follow the generation method proposed by PO3AD \cite{po3ad}, specifically, it divides the point cloud into multiple patches and randomly selects one or more of them. All points within the selected patch are displaced along either the normal direction or its opposite by a certain distance, controlled by a distance scaling factor. This process simulates realistic pseudo-anomalous regions such as protrusions or indentations.

\subsection{Anomaly Score for Inference}

The anomaly level of each point is evaluated based on its predicted classification logits under our point-wise classification fine-tuning. Specifically, for each point, the model outputs a probability distribution over two categories (normal and anomalous) through a softmax activation. Let 
$x^0_i$ and $x^1_i$ denote the predicted probabilities of point $x_i$ belonging to the normal and anomalous classes, respectively. The point-level anomaly score $\phi(x_i)$ is defined as:
\begin{gather}
    \label{eq4}
    \phi(x_i) = -\log\left(\frac{x^0_i}{x^1_i + \varepsilon}\right),
\end{gather}
where $\varepsilon$ is a small constant added for numerical stability. This formulation computes the logarithmic ratio between the normal and abnormal class probabilities, effectively capturing the degree to which a point deviates from normal patterns while enhancing the discriminability and stability of the anomaly score—larger values indicate a higher likelihood of anomaly.

To obtain the sample-level anomaly score $\phi(x)$, we adopt a top-k aggregation strategy, where only the top-k of point-level scores (with the highest anomaly likelihoods) are averaged:

\begin{gather}
    \label{eq5}
    \phi(x) = \frac{1}{k} \sum_{i=1}^{k} \phi(x_i), \quad \text{where } k = \lceil r \cdot N \rceil,
\end{gather}
where $r$ denotes the aggregation rate and $N$ is the total number of points in the sample. This approach emphasizes the most anomalous regions of a point cloud, improving robustness against false positives from scattered noise.

\section{Experiments}

\subsection{Pre-training} 

Since our primary focus is on evaluating the model’s transferability in anomaly detection, we construct our pre-training dataset by collecting the training sets from two classical anomaly detection datasets: Real3D-AD \cite{real3dad} and Anomaly-ShapeNet \cite{anoshapenet}.
Since both training sets consist exclusively of normal samples, our method performs self-supervised learning solely on normal data. Real3D-AD is a high-resolution point cloud dataset collected from real-world objects, covering 12 categories. Each category similarly includes 4 normal samples for training and 100 test instances. Anomaly-ShapeNet is a synthetic dataset derived from ShapeNet \cite{shapenet}, consisting of 1,600 point cloud samples across 40 categories, with each category providing 4 normal training instances. As a result, we obtain a total of 208 distinct normal samples. To enhance diversity, we further apply data augmentation to generate 4 different variants for each sample, resulting in a pre-training dataset comprising 832 normal samples in total.

We train our model for 300 epochs using the Adam optimizer with an initial learning rate of 0.001. The learning rate is decayed by a factor of 0.5 every 10 epochs.

\subsection{Fine-tuning on Downstream Tasks} 

We assess the efficacy of our approach by fine-tuning our pre-trained models on downstream tasks, including anomaly detection, classification, and part segmentation.

\begin{table*}[t]
\resizebox{\textwidth}{!}{
\begin{tabular}{@{}lccccccccccccc|c@{}}
\toprule
\multicolumn{15}{c}{\textbf{(a) O-AUROC($\uparrow$)}}  \\ \midrule
\textbf{Method} & \textbf{Reference}   & \textbf{Airplane} & \textbf{Car} &\textbf{Candybar} & \textbf{Chicken} & \textbf{Diamond} & \textbf{Duck} & \textbf{Fish} & \textbf{Gemstone} & \textbf{Seahorse} & \textbf{Shell} & \textbf{Starfish} & \textbf{Toffees} & \textbf{Average}  \\ \midrule
\textbf{BTF}      & CVPRW'23   & 0.730          & 0.647         & 0.539         & 0.789         & 0.707         & 0.691         & 0.602         & 0.686         & 0.596         & 0.396         & 0.530         & 0.703         & 0.635   \\
\textbf{M3DM}     & CVPR'23    & 0.434          & 0.541         & 0.552         & 0.683         & 0.602         & 0.433         & 0.540         & 0.644         & 0.495         & 0.694         & 0.551         & 0.450         & 0.552   \\
\textbf{PatchCore}& CVPR'22    & \textbf{0.848} & 0.777         & 0.570         & \textbf{0.853}& 0.784         & 0.628         & 0.837         & 0.359         & 0.767         & 0.663         & 0.471         & 0.626         & 0.682   \\
\textbf{CPMF}     & PR'24      & 0.701          & 0.551         & 0.552         & 0.504         & 0.523         & 0.582         & 0.558         & 0.589         & 0.729         & 0.653         & 0.700         & 0.390         & 0.586   \\
\textbf{IMRNet}   & CVPR'24    & 0.762          & 0.711         & 0.755         & 0.780         & 0.905         & 0.517         & 0.880         & 0.674         & 0.604         & 0.665         & 0.674         & 0.774         & 0.725   \\
\textbf{Reg3D-AD} & NeurIPS'23 & 0.716          & 0.697         & 0.827         & 0.852         & 0.900         & 0.584         & 0.915         & 0.417         & 0.762         & 0.583         & 0.506         & 0.685         & 0.704   \\
\textbf{Group3AD} & ACM MM'24  & 0.744          & 0.728         & 0.847         & 0.786         & 0.932         & 0.679         & 0.976         & 0.539         & \textbf{0.841}& 0.585         & 0.562         & 0.796         & 0.751   \\
\textbf{R3D-AD}   & ECCV'24    & 0.772          & 0.696         & 0.713         & 0.714         & 0.685         & \textbf{0.909}& 0.692         & 0.665         & 0.720         & \textbf{0.840}& 0.701         & 0.703         & 0.734   \\
\textbf{PO3AD}    & CVPR'25    & 0.804          & 0.654         & 0.785         & 0.686         & 0.801         & 0.820         & 0.859         & 0.693         & 0.756         & 0.800         & 0.758         & 0.771         & 0.765   \\
\textbf{ISMP}     & AAAI'25    & 0.858          & 0.731         & 0.852         & 0.714         & 0.948         & 0.712         & \textbf{0.945}& 0.468         & 0.729         & 0.623         & 0.660         & 0.842         & 0.767   \\
\rowcolor{mycolor}\textbf{Curvature}& Ours       & 0.345          & 0.686         & \textbf{0.889}& 0.496         & 0.957         & 0.768         & 0.854         & 0.554         & 0.790         & 0.594         & 0.872         & 0.872         & 0.723   \\
\rowcolor{mycolor}\textbf{CASL}  & Ours       & 0.808          & \textbf{0.799}& 0.848         & 0.657         & \textbf{0.976}& 0.836         & 0.935         & \textbf{0.769}& 0.643         & 0.791         & \textbf{0.893}& \textbf{0.924}& \textbf{0.823}   \\
\midrule
& \multicolumn{1}{l}{}                & \multicolumn{1}{l}{}                  & \multicolumn{1}{l}{}                  & \multicolumn{1}{l}{}                  & \multicolumn{1}{l}{}                  & \multicolumn{1}{l}{}                  & \multicolumn{1}{l}{}                  & \multicolumn{1}{l}{}                  & \multicolumn{1}{l}{}                  & \multicolumn{1}{l}{}                  & \multicolumn{1}{l}{}                  & \multicolumn{1}{l}{}                  & \multicolumn{1}{l}{}                  & \multicolumn{1}{l}{} \\ \midrule
\multicolumn{15}{c}{\textbf{(b) P-AUROC($\uparrow$)}}                                                                                                               \\ \midrule
\textbf{Method} & \textbf{Reference}  & \textbf{Airplane} & \textbf{Car} &\textbf{Candybar} & \textbf{Chicken} & \textbf{Diamond} & \textbf{Duck} & \textbf{Fish} & \textbf{Gemstone} & \textbf{Seahorse} & \textbf{Shell} & \textbf{Starfish} & \textbf{Toffees} & \textbf{Average} \\ \midrule
\textbf{BTF}      & CVPRW'23   & 0.738          & 0.708         & 0.864         & 0.693         & 0.882         & 0.875         & 0.709         & 0.891         & 0.512         & 0.571         & 0.501         & 0.815         & 0.722  \\
\textbf{M3DM}     & CVPR'23    & 0.530          & 0.607         & 0.683         & 0.735         & 0.618         & 0.678         & 0.600         & 0.654         & 0.561         & 0.748         & 0.555         & 0.679         & 0.637  \\
\textbf{PatchCore}& CVPR'22    & 0.556          & 0.740         & 0.749         & 0.558         & 0.854         & 0.658         & 0.781         & 0.539         & 0.808         & 0.753         & 0.613         & 0.549         & 0.692  \\
\textbf{Reg3D-AD} & NeurIPS'23 & 0.631          & 0.718         & 0.724         & 0.676         & 0.835         & 0.503         & 0.826         & 0.545         & 0.817         & 0.811         & 0.617         & 0.759         & 0.700  \\
\textbf{Group3AD} & ACM MM'24  & 0.636          & 0.745         & 0.738         & 0.759         & 0.862         & 0.631         & 0.836         & 0.564         & \textbf{0.827}& 0.798         & 0.625         & 0.803         & 0.735  \\
\textbf{ISMP}     & AAAI'25    & 0.753          & 0.836         & 0.907         & \textbf{0.798}& 0.926         & 0.876         & 0.886         & 0.857         & 0.813         & 0.839         & 0.641         & 0.895         & 0.836  \\
\rowcolor{mycolor}\textbf{Curvature}& Ours       & 0.715          & 0.699         & 0.810         & 0.705         & 0.860         & 0.860         & 0.727         & 0.904         & 0.545         & 0.614         & 0.519         & 0.797         & 0.729  \\
\rowcolor{mycolor}\textbf{CASL}  & Ours       & \textbf{0.842} & \textbf{0.905}& \textbf{0.932}& 0.713         & \textbf{0.988}& \textbf{0.895}& \textbf{0.935}& \textbf{0.916}& 0.814         & \textbf{0.873}& \textbf{0.839}& \textbf{0.937}& \textbf{0.882}  \\ 
\bottomrule
\end{tabular}}
\caption{The experimental results for anomaly detection across 12 categories of Real3D-AD. Table (a) reports the object-level AUROC (O-AUROC) scores, while Table (b) presents the results of the point-level AUROC (P-AUROC). The results of the baselines are from their papers.}
  \label{tabreal3d}
\end{table*}

\subsubsection{Anomaly Detection on Real3D-AD.} Our anomaly detection experiments are conducted following the protocols established in previous works \cite{real3dad}. We adopt the Area Under the Receiver Operating Characteristic Curve (AUROC) as our evaluation metric, as it provides an objective measure of both detection (object-level) and localization (point-level) performance without relying on any predefined decision threshold.

The results in the Table \ref{tabreal3d} show the performance of different anomaly detection methods on Real3D-AD. Among them, Curvature is our proposed non-learning-based method that relies purely on curvature to identify anomalies, while CASL is our learning-based method. As shown in the results, our Curvature method outperforms several classical learning-based methods specifically designed for anomaly detection, such as M3DM, CPMF, and Reg3D-AD, demonstrating the importance of curvature in identifying anomalies. Furthermore, our CASL method, which integrates multi-scale curvature prompts, achieves new state-of-the-art (SOTA) performance, surpassing the second-best approach by \textbf{5.6\%} in object-level AUROC and \textbf{4.6\%} in point-level AUROC, highlighting the superiority of our approach.

\subsubsection{Anomaly Detection on Anomaly-ShapeNet.} Table \ref{tabano} reports the experimental results of our anomaly detection approach on Anomaly-ShapeNet, which consists of 40 distinct categories. We present both the O-AUROC and P-AUROC scores for each category, as well as the overall average performance across all categories. Due to space limitations, Table \ref{tabano} only displays the average results; detailed per-category results can be found in the supplementary material. As shown in the table, our Curvature-based method also outperforms several learning-based approaches specifically designed for anomaly detection. Furthermore, our CASL method achieves a new state-of-the-art, surpassing the second-best PO3AD by \textbf{4.8\%} in terms of average O-AUROC, demonstrating the superiority of our approach.

\begin{table}[t]
  \centering
  \resizebox{0.95\linewidth}{!}{
    \begin{tabular}{lccc}
    \toprule
    \textbf{Methods} & \textbf{Reference} & \textbf{Avg. O-AUROC} & \textbf{Avg. P-AUROC} \\
    \midrule
    \textbf{BTF}  & CVPRW'23& 0.528  & 0.628 \\
    \textbf{M3DM} & CVPR'23 & 0.552 & 0.616 \\
    \textbf{PatchCore}  & CVPR'22 & 0.568  & 0.580 \\
    \textbf{CPMF}  & PR'24 & 0.559  & 0.573 \\
    \textbf{Reg3D-AD}  & NeurIPS'23 & 0.572  & 0.668 \\
    \textbf{IMRNet}  & CVPR'24 & 0.661  & 0.650 \\
    \textbf{R3D-AD}  & ECCV'24 & 0.749  & - \\
    \textbf{ISMP}  & AAAI'25 & -  & 0.691 \\
    \textbf{PO3AD}  & CVPR'25 & 0.839  & 0.898 \\
    \rowcolor{mycolor}\textbf{Curvature}  & Ours & 0.559  & 0.693 \\
    \rowcolor{mycolor}\textbf{CASL}  & Ours & \textbf{0.887}  & \textbf{0.899} \\
    \bottomrule
    \end{tabular}%
    }
  \caption{Anomaly detection results on Anomaly-ShapeNet. We report the average O-AUROC and P-AUROC scores across its 40 categories. More detailed results are provided in the supplementary material.}
  \label{tabano}%
\end{table}%

\subsubsection{Object Classification.} We further validate the effectiveness of our model on the classical object classification task. Following prior work \cite{pointbert,pointmae,pointgpt}, we report the classification fine-tuning results of our CASL model on three variants of the real-world ScanObjectNN \cite{scanobjectnn} dataset. ScanObjectNN is a widely used benchmark consisting of approximately 15,000 point cloud samples across 15 categories. These objects represent indoor scenes and are often characterized by cluttered backgrounds and occlusions caused by other objects.

Table \ref{tabclass} presents our experimental results. As shown, our method achieves leading performance on the OBJ-BG and OBJ-ONLY variants. On the more challenging PB-T50-RS variant, our model performs below the best-performing SFR\cite{sfr}. Notably, unlike prior approaches that pretrain on the full ShapeNet \cite{shapenet} dataset (around 50k samples), our model is pretrained on only 832 samples, which is a significantly smaller dataset. Despite this, CASL still matches or exceeds some leading methods, demonstrating the effectiveness of the proposed multi-scale curvature prompts. By leveraging curvature guidance, our method effectively mitigates the geometric shortcut problem and achieves strong representational ability with minimal data.

\begin{table}[t]
  \centering
  \resizebox{\linewidth}{!}{
    \begin{tabular}{lcccc}
    \toprule
    \textbf{Method} & Reference & \textbf{OBJ-BG}($\uparrow$) & \textbf{OBJ-ONLY}($\uparrow$ & \textbf{PB-T50-RS}($\uparrow$) \\
    \midrule
    \multicolumn{5}{c}{\textit{Supervised Learning Only}} \\
    \midrule
    \textbf{PointNet} & CVPR'17   & 73.3  & 79.2  & 68.0 \\
    \textbf{PointNet++} & NeurIPS'17   & 82.3  & 84.3  & 77.9 \\
    \textbf{DGCNN}  & TOG'19 & 82.8  & 86.2  & 78.1 \\
    \textbf{SimpleView} & ICML'21   & -     & -     & 80.8 \\
    \textbf{PointMLP}  & ICLR'22      & -     & -     & 85.7 \\
    \textbf{SFR}    & ICASSP'23  & -     & -     & \textbf{87.8} \\
    \midrule
    \multicolumn{5}{c}{\textit{with Self-supervised Representation Learning}} \\
    \midrule
    \textbf{Point-BERT} & CVPR'22  & 87.43 & 88.12 &  83.07 \\
    \textbf{MaskPoint}  & ECCV'22  & 89.70 & 89.30 &  84.60 \\
    \textbf{Point-MAE}  & ECCV'22   & 90.02 & 88.29 & 85.18 \\
    \textbf{Point-M2AE} & NeurIPS'22   & 91.22 & 88.81 & 86.43 \\
    \textbf{PointGPT-S} & NeurIPS'23   & 91.60 & 90.00 & 86.90 \\
    \rowcolor{mycolor}\textbf{CASL} & Ours  & \textbf{92.08} & \textbf{91.05} & 86.81 \\
    \bottomrule
    \end{tabular}%
    }
  \caption{Classification results on three variants of the ScanObjectNN dataset, and we report the classification accuracy(\%).}
  \label{tabclass}%
\end{table}%

\subsubsection{Part Segmentation.} We further evaluate the performance of CASL on the part segmentation task using the ShapeNetPart dataset, which contains 16,881 samples spanning 16 object categories. For a fair comparison, we adopt the same segmentation setup following previous works \cite{pointmae,pointgpt}. Table \ref{seg} reports our experimental results. Similarly, despite pretraining on a significantly smaller dataset, our method achieves equally good performance in the part segmentation task.

\begin{table}[t]
  \centering
  \resizebox{0.7\linewidth}{!}{
    \begin{tabular}{lcll}
    \toprule
    \textbf{Methods} & \textbf{Reference} & \textbf{$\mathrm{mIoU}_{c}$} & \textbf{$\mathrm{mIoU}_{I}$} \\
    \midrule
    \multicolumn{4}{c}{\textit{Supervised Learning Only}} \\
    \midrule
    \textbf{PointNet++}  & NeurIPS'17& 81.9  & 85.1 \\
    \textbf{DGCNN} & TOG'19 & 82.3  & 85.2 \\
    \midrule
    \multicolumn{4}{c}{\textit{with Self-supervised Representation Learning}} \\
    \midrule
    \textbf{PointContrast}  & ECCV'20 & -  & 85.1 \\
    \textbf{IDPT}  & ICCV'23 & 83.8  & 85.9 \\
    \textbf{Point-BERT}  & CVPR'22 & 84.1  & 85.6 \\
    \textbf{Point-MAE}  & ECCV'22 & \textbf{84.2}  & 86.1 \\
    \textbf{PointGPT-S}  & NeurIPS'23 & 84.1  & 86.2 \\
    \rowcolor{mycolor}\textbf{CASL}  & Ours & 84.1  & \textbf{86.2} \\
    \bottomrule
    \end{tabular}%
    }
  \caption{Part segmentation results on the ShapeNetPart. The mean IoU across all categories, i.e., $\mathrm{mIoU}_{c}$ (\%), and the mean IoU across all instances, i.e., $\mathrm{mIoU}_{I}$ (\%) are reported.}
  \label{seg}%
\end{table}%

\subsection{Ablation Study} 

\subsubsection{The Effect of Masked Points and Curvature Prompts.}
We further analyze the two important components of our method in Table \ref{abl1}, namely the masked points and the curvature prompts.
Experiment A performs reconstruction from full original coordinates back to themselves using only the coordinate encoder, without applying any masking.
Experiment B also relies solely on the coordinate encoder to reconstruct coordinates, but with a 60\% random masking ratio applied during encoding.
Experiment C uses only the curvature encoder without any masking, aiming to reconstruct coordinates from the curvature space.
Experiment D represents our proposed method, which reconstructs coordinates using curvature prompts and the coordinate encoder with fully masked points (100\% masking ratio). 

As shown in Table \ref{abl1}, A achieves the lowest reconstruction loss, but detection results are very poor. Because it performs self-reconstruction from the coordinate to itself, which introduces significant geometric shortcuts, ultimately leading to representation collapse.
B mitigates this issue to some extent by randomly masking a large portion of the original coordinate space and reconstructing from the masked points to the original domain. However, the learned representations remain suboptimal.
Experiment C learns to reconstruct coordinates from the curvature space, resulting in substantial improvements over A and B. Our method D further incorporates curvature prompts into the decoder and operates in a fully masked coordinate space, substantially reducing geometric shortcuts and thereby enabling superior representational learning.

\begin{table}[t]
  \centering
  \resizebox{\linewidth}{!}{
    \begin{tabular}{cccccc}
    \toprule
     & Masked Points & Curvature Prompts & O-AUROC & P-AUROC & Rec. Loss \\
    \midrule
    A & \ding{56}     & \ding{56}     & 0.751 & 0.649 & 0.008\\
    B & \ding{52}     & \ding{56}     & 0.776 & 0.711 & 0.012\\
    C & \ding{56}     & \ding{52}     & 0.804 & 0.878 & 0.019\\
    D & \ding{52}     & \ding{52}     & 0.823 & 0.882 & 0.016\\
    \bottomrule
    \end{tabular}%
  }
  \caption{The effect of masked points and curvature prompts.}
  \label{abl1}%
\end{table}%

\begin{figure}[t]
    \begin{center}
    \includegraphics[width=\linewidth]{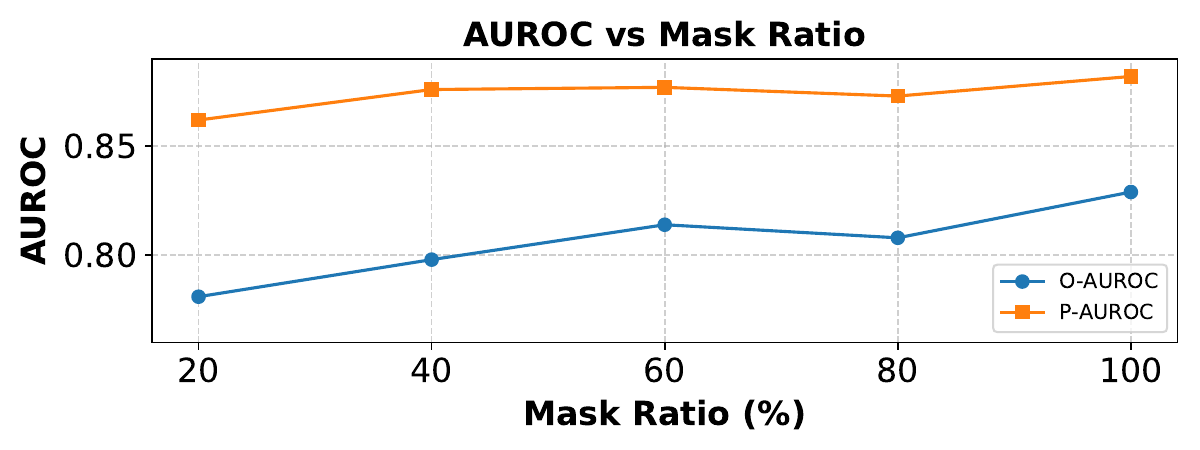}
    \caption{Detection performance with varying mask ratios.
    }\label{rate}
    \end{center}
\end{figure}

\subsubsection{The Effect of Masked Ratios.} We investigate the effect of different masking ratios in our coordinate encoder on anomaly detection. As shown in Figure \ref{rate}, the best performance is achieved at 100\% masking ratio, indicating that any additional geometric information in the coordinate space during curvature-guided reconstruction introduces geometric shortcuts, thereby limiting the representation learning.

\section{Conclusions}

In this paper, we demonstrate that classical pre-trained point cloud models perform suboptimally in anomaly detection under a unified fine-tuning paradigm due to geometric shortcuts. To address this, we propose CASL, a curvature-augmented reconstruction framework that leverages multi-scale curvature prompts and full-coordinate masking to guide learning. CASL effectively mitigates geometric shortcuts and enhances 3D structural understanding. Experiments on Real3D-AD and Anomaly-ShapeNet show that CASL achieves leading anomaly detection performance and generalizes well to other 3D understanding tasks.

\section{Acknowledgments}

This work is supported in part by the National Natural Science Foundation of China, under Grant (62302309,62571298), Shenzhen Science and Technology Program (JCYJ20220818101014030).

\bibliography{aaai2026}

\clearpage

\section{Supplementary Materials}

\subsection{Detailed Structure of the Encoder Block}

\begin{figure}[t]
    \begin{center}
    \includegraphics[width=\linewidth]{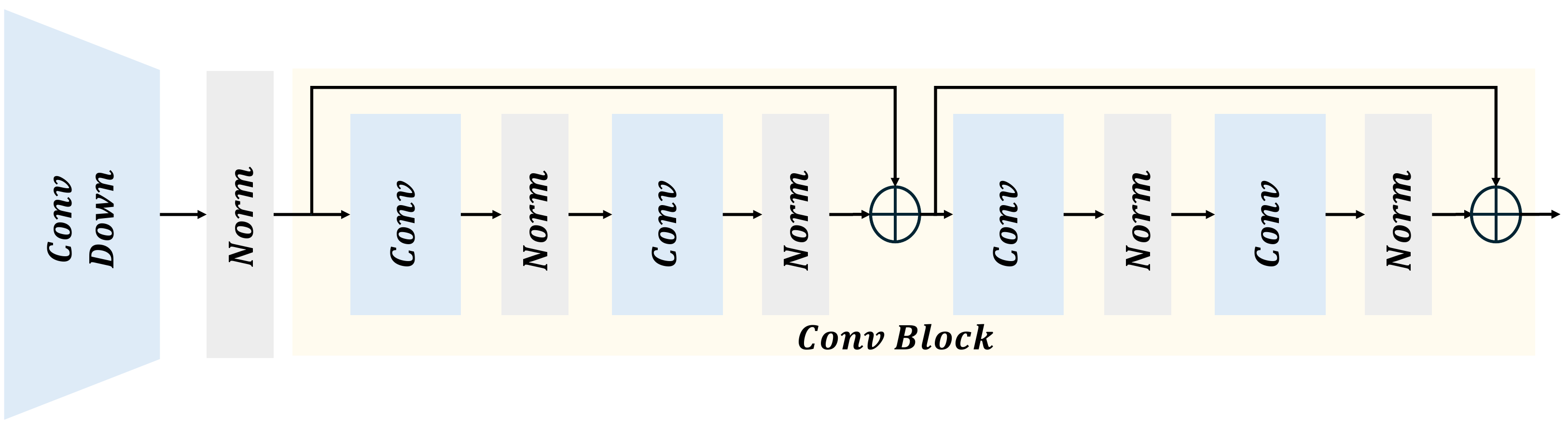}
    \caption{Detailed structure of the encoder block.
    }\label{block}
    \end{center}
\end{figure}

The encoder block, as illustrated in Figure \ref{block}, begins with a downsampling convolution layer (Conv Down), which reduces the spatial resolution of the input while mapping its feature dimensionality. This is followed by a normalization layer (Norm) to stabilize the learning process and improve convergence.

The core component of the encoder is the Convolutional Block (Conv Block), which is composed of four sequential convolutional layers, each with a stride of 1, designed to preserve spatial dimensions while progressively refining feature representations. Each convolution layer is followed by a normalization layer to maintain feature stability throughout the block.

To enhance feature propagation and mitigate the vanishing gradient problem, residual connections are incorporated after every two convolutional operations. Specifically, the output of the first convolution is element-wise added to the output of the third convolution, and similarly, the output of the third is added to the output of the final (fifth) convolution. These residual connections enable the block to learn identity mappings more easily, facilitating deeper and more expressive feature learning.

This encoder block structure effectively captures both low-level and mid-level geometric features in the input data, serving as a strong backbone for subsequent high-level processing.

\subsection{More Curvature and Abnormal Visualization}

\begin{figure*}[ht!]
    \begin{center}
    \includegraphics[width=\linewidth]{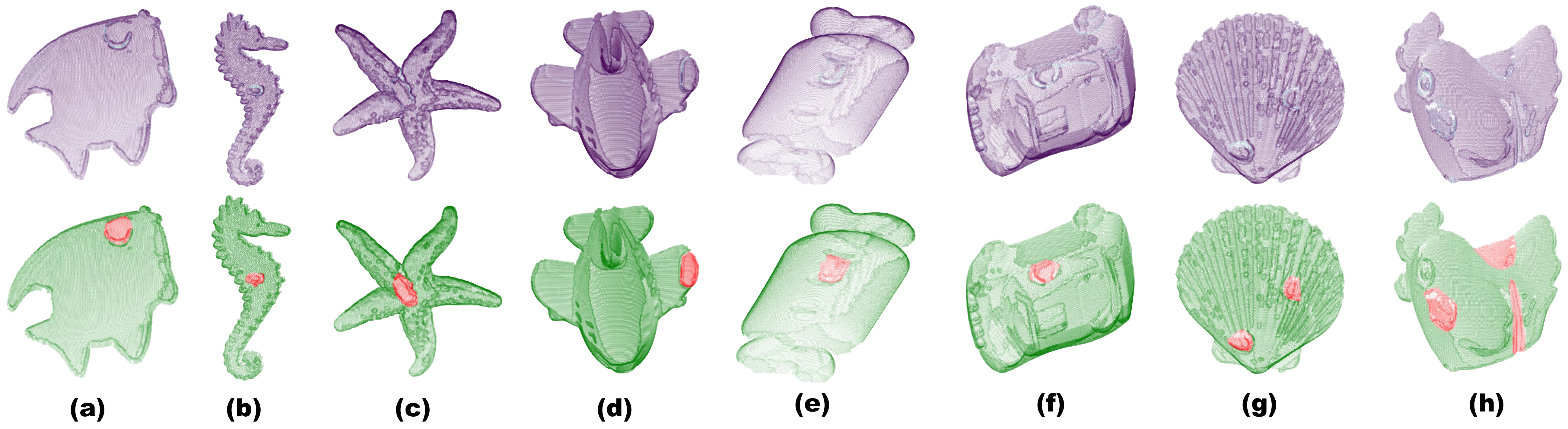}
    \caption{Comparisons of curvature heatmaps (top) and anomaly distributions (bottom). Warmer colors in the heatmaps indicate higher curvature. In the anomaly distribution maps, anomalous points are shown in red, and normal points in green.
    }\label{heatmap_supp}
    \end{center}
\end{figure*}

Figure \ref{heatmap_supp} presents a visual comparison of curvature heatmaps (top) and anomaly maps (bottom) across several representative samples. In the curvature heatmaps, warmer colors indicate regions with higher curvature, which often correspond to sharp geometric structures or surface discontinuities. These regions typically exhibit richer local geometric information and are more likely to be indicative of structural abnormalities.

In the anomaly visualizations, red points denote those identified as anomalous, while green points represent normal regions. As shown in the figure, our method accurately highlights local geometric irregularities, especially around edges, corners, or concave regions—many of which correlate strongly with areas of high curvature. This alignment validates the effectiveness of incorporating curvature priors during pretraining, as it encourages the model to focus on semantically meaningful geometric variations.

\subsection{Reconstruction Visualization of CASL}

\begin{figure*}[ht!]
    \begin{center}
    \includegraphics[width=\linewidth]{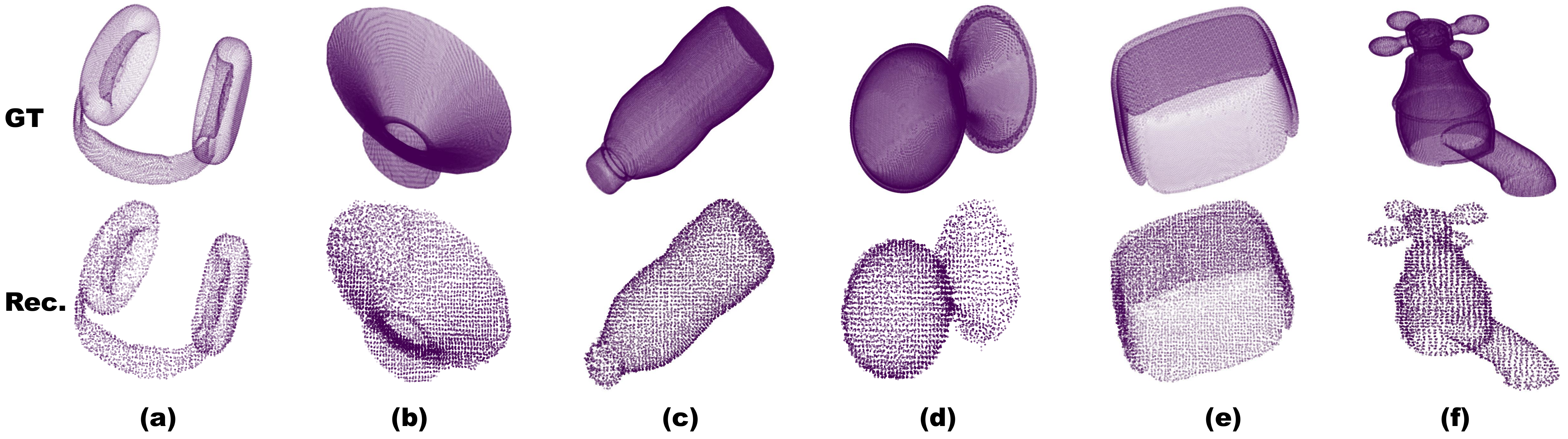}
    \caption{Comparison between the ground truth (top row, GT) and the reconstructed point clouds (bottom row, Rec.) generated by our method guided solely by curvature features. Despite the absence of explicit spatial coordinates, our model successfully restores fine-grained geometric structures across diverse object categories. Subfigures (a)--(f) represent representative classes such as headphones, dishes, bottles, and faucets.
    }\label{rec_supp}
    \end{center}
\end{figure*}

Figure \ref{rec_supp} presents qualitative results of our reconstruction method CASL, which leverages only local curvature features to reconstruct the complete point cloud geometry. The upper row shows the ground truth (GT) shapes, while the lower row displays the corresponding reconstructions (Rec.) generated by our curvature-guided decoder.

Remarkably, despite receiving no direct spatial coordinates as input, the model is capable of capturing both global object structure and local surface details. For example, in (a) and (b), the contour and depth of curved surfaces are faithfully preserved. Even in more complex shapes such as bottles (c), symmetric discs (d), and topologically intricate objects like faucets (f), our method produces reconstructions that are topologically and geometrically consistent with the original data. These results strongly support the hypothesis that curvature encodes rich structural information, and that our model can effectively utilize such cues to recover accurate 3D geometry, even under highly under-constrained conditions.

\subsection{Qualitative Visualization of CASL}

\begin{figure*}[ht!]
    \begin{center}
    \includegraphics[width=\linewidth]{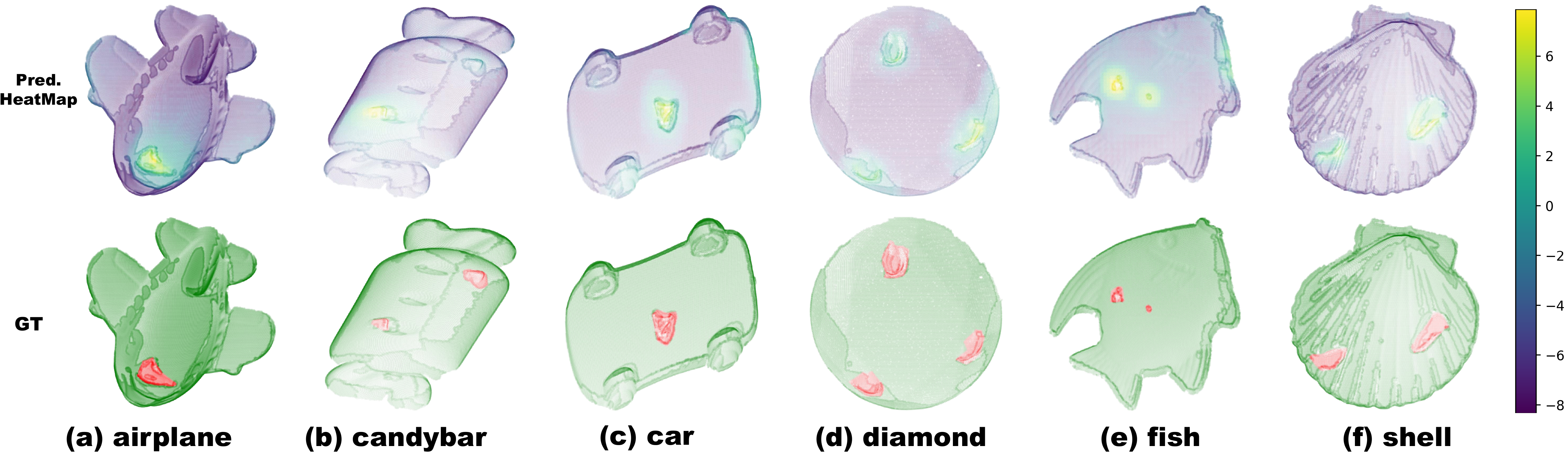}
    \caption{Qualitative visualization of anomaly detection results. The top row shows the predicted heatmaps of anomalies, where warmer colors (e.g., red/yellow) indicate higher anomaly scores. The bottom row illustrates the ground truth (GT) of the anomalies.
    }\label{anopred_supp}
    \end{center}
\end{figure*}

Figure \ref{anopred_supp} presents some qualitative results of our method on the Real3D-AD dataset, highlighting the predicted anomaly heatmaps (top) and the corresponding ground truth annotations (bottom). As demonstrated across various object categories, our model successfully identifies the anomalous regions with high precision. Notably, even in samples containing multiple or spatially dispersed anomalies—such as the diamond and shell categories—our method produces consistent and accurate predictions. These visual results further validate the effectiveness of our approach in precisely localizing anomalies beyond simple cases, showcasing its robustness under complex anomaly scenarios.

\subsection{Detailed Results on the Anomaly-ShapeNet Dataset} 

Anomaly-ShapeNet \cite{anoshapenet} is a synthetic dataset constructed from the ShapeNet repository \cite{shapenet}, comprising a total of 1,600 point cloud samples uniformly distributed across 40 object categories. This dataset is widely used to benchmark point cloud anomaly detection methods due to its diversity in object shapes and fine-grained structural variations.

In the main paper, we reported the average anomaly detection performance over all 40 categories. To provide a more comprehensive evaluation, we present here the category-wise results for both our method and several representative baseline approaches.

Specifically, Table \ref{tabanoo} reports the object-level AUCROC (O-AUROC) scores for each individual category, while Table \ref{tabanop} shows the corresponding point-level AUCROC (P-AUROC) results. Our method consistently achieves state-of-the-art average performance across all categories in both metrics. Furthermore, it outperforms existing methods in the vast majority of individual categories, demonstrating its strong capability in capturing both global and local geometric anomalies.

\begin{table*}[ht]
  \centering
\resizebox{\textwidth}{!}{
    \begin{tabular}{l|cccccccccccccc}
    \toprule
    \multicolumn{15}{c}{\textbf{O-AUROC($\uparrow$)}} \\
    \midrule
    \textbf{Method} & \textbf{cap0} & \textbf{cap3} & \textbf{helmet3} & \textbf{cup0} & \textbf{bowl4} & \textbf{vase3} & \textbf{headset1} & \textbf{eraser0} & \textbf{vase8} & \textbf{cap4} & \textbf{vase2} & \textbf{vase4} & \textbf{helmet0} & \textbf{bucket1} \\
    \midrule
\textbf{BTF}            & 0.618         & 0.522         & 0.444         & 0.586         & 0.609         & 0.699         & 0.490         & 0.719         & 0.668         & 0.520         & 0.546         & 0.510         & 0.571         & 0.633  \\
\textbf{M3DM}           & 0.557         & 0.423         & 0.374         & 0.539         & 0.464         & 0.439         & 0.617         & 0.627         & 0.663         & \textbf{0.777}& 0.737         & 0.476         & 0.526         & 0.501  \\
\textbf{Patchcore}      & 0.580         & 0.453         & 0.404         & 0.600         & 0.494         & 0.449         & 0.637         & 0.657         & 0.662         & 0.757         & 0.721         & 0.506         & 0.546         & 0.551  \\
\textbf{CPMF}           & 0.601         & 0.551         & 0.520         & 0.497         & 0.683         & 0.582         & 0.458         & 0.689         & 0.529         & 0.553         & 0.582         & 0.514         & 0.555         & 0.601  \\
\textbf{Reg3D-AD}       & 0.693         & 0.725         & 0.367         & 0.510         & 0.663         & 0.650         & 0.610         & 0.343         & 0.620         & 0.643         & 0.605         & 0.500         & 0.600         & 0.752  \\
\textbf{IMRNet}         & 0.737         & 0.775         & 0.573         & 0.643         & 0.676         & 0.700         & 0.676         & 0.548         & 0.630         & 0.652         & 0.614         & 0.524         & 0.597         & 0.771  \\
\textbf{R3D-AD}         & 0.822         & 0.730         & 0.707         & 0.776         & 0.744         & 0.742         & 0.795         & 0.890         & 0.721         & 0.681         & 0.752         & 0.630         & 0.757         & 0.756  \\
\textbf{PO3AD}          & 0.877         & \textbf{0.859}& 0.754         & 0.871         & \textbf{0.981}& 0.821         & 0.923         & 0.995         & 0.739         & 0.792         & 0.952         & 0.675         & 0.762         & 0.787  \\
\rowcolor{mycolor}\textbf{Curvature}      & 0.663         & 0.611         & 0.461         & 0.633         & 0.500         & 0.579         & 0.705         & 0.710         & 0.542         & 0.386         & 0.586         & 0.518         & 0.522         & 0.587  \\
\rowcolor{mycolor}\textbf{CASL}        & \textbf{0.948}& 0.821         & \textbf{0.861}& \textbf{0.990}& 0.944         & \textbf{0.930}& \textbf{0.967}& \textbf{0.995}& \textbf{0.933}& 0.772         & \textbf{0.981}& \textbf{0.821}& \textbf{0.783}& \textbf{0.914}  \\
    \midrule
    \multicolumn{1}{c}{} &       &       &       &       &       &       &       &       &       &       &       &       &       &  \\
    \midrule
    \textbf{Method} & \textbf{bottle3} & \textbf{vase0} & \textbf{bottle0} & \textbf{tap1} & \textbf{bowl0} & \textbf{bucket0} & \textbf{vase5} & \textbf{vase1} & \textbf{vase9} & \textbf{ashtray0} & \textbf{bottle1} & \textbf{tap0} & \textbf{phone0} & \textbf{cup1} \\
    \midrule
\textbf{BTF}            & 0.322         & 0.342         & 0.344         & 0.546         & 0.509         & 0.401         & 0.409         & 0.219         & 0.268         & 0.420         & 0.546         & 0.560         & 0.671         & 0.610  \\
\textbf{M3DM}           & 0.541         & 0.423         & 0.574         & 0.739         & 0.634         & 0.309         & 0.317         & 0.427         & 0.663         & 0.577         & 0.637         & 0.754         & 0.357         & 0.556  \\
\textbf{Patchcore}      & 0.572         & 0.455         & 0.604         & 0.766         & 0.504         & 0.469         & 0.417         & 0.423         & 0.660         & 0.587         & 0.667         & 0.753         & 0.388         & 0.586  \\
\textbf{CPMF}           & 0.405         & 0.451         & 0.520         & 0.697         & 0.783         & 0.482         & 0.618         & 0.345         & 0.609         & 0.353         & 0.482         & 0.359         & 0.509         & 0.499  \\
\textbf{Reg3D-AD}       & 0.525         & 0.533         & 0.486         & 0.641         & 0.671         & 0.610         & 0.520         & 0.702         & 0.594         & 0.597         & 0.695         & 0.676         & 0.414         & 0.538  \\
\textbf{IMRNet}         & 0.640         & 0.533         & 0.552         & 0.696         & 0.681         & 0.580         & 0.676         & 0.757         & 0.594         & 0.671         & 0.700         & 0.676         & 0.755         & 0.757  \\
\textbf{R3D-AD}         & 0.781         & 0.788         & 0.733         & \textbf{0.900}& 0.819         & 0.683         & 0.757         & 0.729         & 0.718         & 0.833         & 0.737         & 0.736         & 0.762         & 0.757 \\
\textbf{PO3AD}          & 0.926         & \textbf{0.858}& 0.900         & 0.681         & 0.922         & 0.853         & \textbf{0.852}& 0.742         & 0.830         & \textbf{1.000}& 0.933         & 0.745         & 0.776         & \textbf{0.833} \\
\rowcolor{mycolor}\textbf{Curvature}      & 0.743         & 0.500         & 0.681         & 0.559         & 0.600         & 0.638         & 0.362         & 0.557         & 0.524         & 0.362         & 0.670         & 0.630         & 0.729         & 0.476 \\
\rowcolor{mycolor}\textbf{CASL}        & \textbf{1.000}& 0.838         & \textbf{0.957}& 0.563         & \textbf{1.000}& \textbf{0.990}& 0.714         & \textbf{0.890}& \textbf{0.818}& 0.943         & \textbf{0.954}& \textbf{0.761}& \textbf{0.981}& 0.648 \\
    \midrule
    \multicolumn{1}{c}{} &       &       &       &       &       &       &       &       &       &       &       &       &       &  \\
    \midrule
    \textbf{Method} & \textbf{vase7} & \textbf{helmet2} & \textbf{cap5} & \textbf{shelf0} & \textbf{bowl5} & \textbf{bowl3} & \textbf{helmet1} & \textbf{bowl1} & \textbf{headset0} & \textbf{bag0} & \textbf{bowl2} & \textbf{jar0} & \multicolumn{2}{|c}{\textbf{Mean}} \\
    \midrule
\textbf{BTF}           & 0.518          & 0.542         & 0.586         & 0.609         & 0.699         & 0.490         & 0.719         & 0.668         & 0.520         & 0.546         & 0.510         & \multicolumn{1}{c|}{0.424}   & \multicolumn{2}{c}{0.528}   \\
\textbf{M3DM}          & 0.657          & 0.623         & 0.639         & 0.564         & 0.409         & 0.617         & 0.427         & 0.663         & 0.577         & 0.537         & 0.684         & \multicolumn{1}{c|}{0.441}   & \multicolumn{2}{c}{0.552}   \\
\textbf{Patchcore}     & 0.693          & 0.425         & \textbf{0.790}& 0.494         & 0.558         & 0.537         & 0.484         & 0.639         & 0.583         & 0.571         & 0.615         & \multicolumn{1}{c|}{0.472}   & \multicolumn{2}{c}{0.568}   \\
\textbf{CPMF}          & 0.397          & 0.462         & 0.697         & 0.685         & 0.685         & 0.658         & 0.589         & 0.639         & 0.643         & 0.643         & 0.625         & \multicolumn{1}{c|}{0.610}   & \multicolumn{2}{c}{0.559}    \\
\textbf{Reg3D-AD}      & 0.462          & 0.614         & 0.467         & 0.688         & 0.593         & 0.348         & 0.381         & 0.525         & 0.537         & 0.706         & 0.490         & \multicolumn{1}{c|}{0.592}   & \multicolumn{2}{c}{0.572}    \\
\textbf{IMRNet}        & 0.635          & 0.641         & 0.652         & 0.603         & 0.710         & 0.599         & 0.600         & 0.702         & 0.720         & 0.660         & 0.685         & \multicolumn{1}{c|}{0.780}   & \multicolumn{2}{c}{0.661}    \\
\textbf{R3D-AD}        & 0.771          & 0.633         & 0.670         & 0.696         & 0.656         & 0.767         & 0.720         & 0.778         & 0.738         & 0.720         & 0.741         & \multicolumn{1}{c|}{0.838}   & \multicolumn{2}{c}{0.749}    \\
\textbf{PO3AD}         & 0.966          & 0.869         & 0.670         & 0.573         & 0.849         & 0.881         & \textbf{0.961}& 0.829         & 0.808         & 0.833         & 0.833         & \multicolumn{1}{c|}{0.866}   & \multicolumn{2}{c}{0.839}    \\
\rowcolor{mycolor}\textbf{Curvature}     & 0.733          & 0.437         & 0.368         & 0.545         & 0.442         & 0.737         & 0.538         & 0.467         & 0.436         & 0.576         & 0.433         & \multicolumn{1}{c|}{0.629}   & \multicolumn{2}{c}{0.559}    \\
\rowcolor{mycolor}\textbf{CASL}       & \textbf{1.000} & \textbf{0.881}& 0.554         & \textbf{0.832}& \textbf{0.888}& \textbf{0.996}& 0.890         & \textbf{0.933}& \textbf{0.836}& \textbf{0.948}& \textbf{0.993}& \multicolumn{1}{c|}{\textbf{0.995}}   & \multicolumn{2}{c}{\textbf{0.887}}    \\
    \bottomrule
    \end{tabular}%
}
  \caption{The object-level AUROC experimental results for anomaly detection across 40 categories of Anomaly-ShapeNet.}
  \label{tabanoo}%
\end{table*}


\begin{table*}[t!]
  \centering
\resizebox{\textwidth}{!}{
    \begin{tabular}{l|cccccccccccccc}
    \toprule
    \multicolumn{15}{c}{\textbf{P-AUROC($\uparrow$)}} \\
    \midrule
    \textbf{Method} & \textbf{cap0} & \textbf{cap3} & \textbf{helmet3} & \textbf{cup0} & \textbf{bowl4} & \textbf{vase3} & \textbf{headset1} & \textbf{eraser0} & \textbf{vase8} & \textbf{cap4} & \textbf{vase2} & \textbf{vase4} & \textbf{helmet0} & \textbf{bucket1} \\
    \midrule
\textbf{BTF}            & 0.730         & 0.658         & 0.724         & 0.790         & 0.679         & 0.699         & 0.591         & 0.719         & 0.662         & 0.524         & 0.646         & 0.710         & 0.575         & 0.633 \\
\textbf{M3DM}           & 0.531         & 0.605         & 0.655         & 0.715         & 0.624         & 0.658         & 0.585         & 0.710         & 0.551         & 0.718         & 0.737         & 0.655         & 0.599         & 0.699 \\
\textbf{Patchcore}      & 0.472         & 0.653         & 0.737         & 0.655         & 0.720         & 0.430         & 0.464         & 0.810         & 0.575         & 0.595         & 0.721         & 0.505         & 0.548         & 0.571 \\
\textbf{CPMF}           & 0.601         & 0.551         & 0.520         & 0.497         & 0.683         & 0.582         & 0.458         & 0.689         & 0.529         & 0.553         & 0.582         & 0.514         & 0.555         & 0.601 \\
\textbf{Reg3D-AD}       & 0.632         & 0.718         & 0.620         & 0.685         & 0.800         & 0.511         & 0.626         & 0.755         & 0.811         & 0.815         & 0.405         & 0.755         & 0.600         & 0.725 \\
\textbf{IMRNet}         & 0.715         & 0.706         & 0.663         & 0.643         & 0.576         & 0.401         & 0.476         & 0.548         & 0.635         & 0.753         & 0.614         & 0.524         & 0.598         & 0.774 \\
\textbf{ISMP}           & 0.865         & 0.734         & 0.722         & 0.869         & 0.740         & 0.762         & 0.702         & 0.706         & 0.851         & 0.753         & 0.733         & 0.545         & 0.683         & 0.672 \\
\textbf{PO3AD}          & 0.957         & 0.948         & 0.846         & 0.909         & 0.967         & 0.884         & 0.907         & \textbf{0.974}& 0.950         & 0.940         & \textbf{0.978}& 0.902         & 0.878         & 0.899 \\
\rowcolor{mycolor}\textbf{Curvature}      & 0.767         & 0.658         & 0.713         & 0.827         & 0.743         & 0.723         & 0.636         & 0.806         & 0.911         & 0.622         & 0.765         & 0.627         & 0.662         & 0.791 \\
\rowcolor{mycolor}\textbf{CASL}        & \textbf{0.974}& \textbf{0.971}& \textbf{0.980}& \textbf{0.980}& \textbf{0.996}& \textbf{0.931}& \textbf{0.909}& 0.876         & \textbf{0.989}& \textbf{0.949}& 0.971         & \textbf{0.913}& \textbf{0.923}& \textbf{0.927} \\
    \midrule
    \multicolumn{1}{c}{} &       &       &       &       &       &       &       &       &       &       &       &       &       &  \\
    \midrule
    \textbf{Method} & \textbf{bottle3} & \textbf{vase0} & \textbf{bottle0} & \textbf{tap1} & \textbf{bowl0} & \textbf{bucket0} & \textbf{vase5} & \textbf{vase1} & \textbf{vase9} & \textbf{ashtray0} & \textbf{bottle1} & \textbf{tap0} & \textbf{phone0} & \textbf{cup1} \\
    \midrule
\textbf{BTF}            & 0.622         & 0.642         & 0.641         & 0.596         & 0.710         & 0.401         & 0.429         & 0.619         & 0.568         & 0.624         & 0.549         & 0.568         & 0.675         & 0.619 \\
\textbf{M3DM}           & 0.532         & 0.608         & 0.663         & 0.712         & 0.658         & 0.698         & 0.642         & 0.602         & 0.663         & 0.577         & 0.637         & 0.654         & 0.358         & 0.556 \\
\textbf{Patchcore}      & 0.512         & 0.655         & 0.654         & \textbf{0.768}& 0.524         & 0.459         & 0.447         & 0.453         & 0.663         & 0.597         & 0.687         & 0.733         & 0.488         & 0.596 \\
\textbf{CPMF}           & 0.435         & 0.458         & 0.521         & 0.657         & 0.745         & 0.486         & 0.651         & 0.486         & 0.545         & 0.615         & 0.571         & 0.458         & 0.545         & 0.509 \\
\textbf{RegAD}          & 0.525         & 0.548         & 0.888         & 0.741         & 0.775         & 0.619         & 0.624         & 0.602         & 0.694         & 0.698         & 0.696         & 0.589         & 0.599         & 0.698 \\
\textbf{IMRNet}         & 0.641         & 0.535         & 0.556         & 0.699         & 0.781         & 0.585         & 0.682         & 0.685         & 0.691         & 0.671         & 0.702         & 0.681         & 0.742         & 0.688 \\
\textbf{ISMP}           & 0.775         & 0.661         & 0.770         & 0.552         & 0.851         & 0.524         & 0.472         & 0.843         & 0.615         & 0.603         & 0.568         & 0.522         & 0.661         & 0.600 \\
\textbf{PO3AD}          & 0.880         & \textbf{0.955}& \textbf{0.912}& 0.692         & \textbf{0.978}& 0.755         & \textbf{0.937}& 0.882         & \textbf{0.952}& \textbf{0.962}& 0.844         & \textbf{0.783}& 0.810         & \textbf{0.932} \\
\rowcolor{mycolor}\textbf{Curvature}      & 0.814         & 0.580         & 0.717         & 0.519         & 0.823         & 0.567         & 0.495         & 0.702         & 0.612         & 0.473         & 0.663         & 0.565         & 0.771         & 0.456 \\
\rowcolor{mycolor}\textbf{CASL}        & \textbf{0.914}& 0.894         & 0.853         & 0.602         & 0.958         & \textbf{0.796}& 0.732         & \textbf{0.972}& 0.886         & 0.887         & \textbf{0.862}& 0.652         & \textbf{0.838}& 0.797 \\
    \midrule
    \multicolumn{1}{c}{} &       &       &       &       &       &       &       &       &       &       &       &       &       &  \\
    \midrule
    \textbf{Method} & \textbf{vase7} & \textbf{helmet2} & \textbf{cap5} & \textbf{shelf0} & \textbf{bowl5} & \textbf{bowl3} & \textbf{helmet1} & \textbf{bowl1} & \textbf{headset0} & \textbf{bag0} & \textbf{bowl2} & \textbf{jar0} & \multicolumn{2}{|c}{\textbf{Mean}} \\
    \midrule
\textbf{BTF}            & 0.540         & 0.643         & 0.586         & 0.619         & 0.699         & 0.690         & 0.749         & 0.768         & 0.620         & 0.746         & 0.518         & \multicolumn{1}{c|}{0.427} & \multicolumn{2}{c}{0.628} \\
\textbf{M3DM}           & 0.517         & 0.623         & 0.655         & 0.554         & 0.489         & 0.657         & 0.427         & 0.663         & 0.581         & 0.637         & 0.694         & \multicolumn{1}{c|}{0.541} & \multicolumn{2}{c}{0.616} \\
\textbf{Patchcore}      & 0.693         & 0.455         & 0.795         & 0.613         & 0.358         & 0.327         & 0.489         & 0.531         & 0.583         & 0.574         & 0.625         & \multicolumn{1}{c|}{0.478} & \multicolumn{2}{c}{0.580} \\
\textbf{CPMF}           & 0.504         & 0.515         & 0.551         & 0.783         & 0.684         & 0.641         & 0.542         & 0.488         & 0.699         & 0.655         & 0.635         & \multicolumn{1}{c|}{0.611} & \multicolumn{2}{c}{0.573} \\
\textbf{RegAD}          & 0.881         & 0.825         & 0.467         & 0.688         & 0.691         & 0.654         & 0.624         & 0.645         & 0.580         & 0.715         & 0.593         & \multicolumn{1}{c|}{0.599} & \multicolumn{2}{c}{0.668} \\
\textbf{IMRNet}         & 0.593         & 0.644         & 0.742         & 0.605         & 0.715         & 0.599         & 0.604         & 0.705         & 0.615         & 0.668         & 0.684         & \multicolumn{1}{c|}{0.765} & \multicolumn{2}{c}{0.650} \\
\textbf{ISMP}           & 0.701         & 0.844         & 0.678         & 0.687         & 0.534         & 0.773         & 0.622         & 0.546         & 0.580         & 0.747         & 0.736         & \multicolumn{1}{c|}{0.823} & \multicolumn{2}{c}{0.691} \\
\textbf{PO3AD}          & 0.982         & \textbf{0.932}& \textbf{0.864}& 0.663         & 0.941         & 0.935         & \textbf{0.948}& 0.914         & \textbf{0.823}& 0.949         & 0.918         & \multicolumn{1}{c|}{0.871} & \multicolumn{2}{c}{0.898} \\
\rowcolor{mycolor}\textbf{Curvature}      & 0.823         & 0.861         & 0.630         & 0.696         & 0.592         & 0.857         & 0.492         & 0.702         & 0.565         & 0.869         & 0.827         & \multicolumn{1}{c|}{0.813} & \multicolumn{2}{c}{0.693} \\
\rowcolor{mycolor}\textbf{CASL}        & \textbf{0.997}& 0.929         & 0.669         & \textbf{0.928}& \textbf{0.974}& \textbf{0.998}& 0.882         & \textbf{0.975}& 0.770         & \textbf{0.968}& \textbf{0.982}& \multicolumn{1}{c|}{\textbf{0.974}} & \multicolumn{2}{c}{\textbf{0.899}} \\
    \bottomrule
    \end{tabular}%
}
  \caption{The point-level AUROC experimental results for anomaly detection across 40 categories of Anomaly-ShapeNet.}
  \label{tabanop}%
\end{table*}

\end{document}